\documentclass{article}

\PassOptionsToPackage{numbers}{natbib}

\usepackage[preprint]{neurips_2024}

\usepackage[utf8]{inputenc} 
\usepackage[T1]{fontenc}    
\usepackage{hyperref}       
\usepackage{url}            
\usepackage{booktabs}       
\usepackage{amsfonts}       
\usepackage{nicefrac}       
\usepackage{microtype}      
\usepackage{xcolor}         

\usepackage{amsmath}
\usepackage{multirow}
\usepackage{siunitx}
\usepackage{graphicx}
\usepackage{colortbl}
\usepackage{pgf-pie}
\usepackage{algorithm}
\usepackage{algpseudocode}
\usepackage{subcaption} 

\definecolor{CC_GREEN}{HTML}{4DB24D}
\definecolor{lightgray1}{rgb}{0.9, 0.9, 0.9} 
\definecolor{customPurple}{RGB}{155, 89, 182}

\title{EXAQ: Exponent Aware Quantization For LLMs Acceleration}

\author{
Moran Shkolnik\,${^\dagger}{^\circ}$\thanks{Equal contribution}\quad
Maxim Fishman\,$^\dagger$\footnotemark[1]\quad
Brian Chmiel\,${^\dagger}$\quad
\\[0.15cm]\textbf{
Hilla Ben-Yaacov\,$^\dagger$\quad
Ron Banner\,$^\dagger$\quad
Kfir Yehuda Levy\,$^\circ$\quad}
\\[0.2cm]
$^\dagger$Habana Labs  --  An Intel company, Caesarea, Israel,\\
$^\circ$Department of Electrical Engineering - Technion, Haifa, Israel
\\[0.2cm]
\small{\texttt{\{\href{mailto:mshkolnik@habana.ai}{mshkolnik},
\href{mailto:mfishman@habana.ai}{mfishman},
\href{mailto:bchmiel@habana.ai}{bchmiel},
\href{mailto:hbyaacov@habana.ai}{hbyaacov},
\href{mailto:rbanner@habana.ai}{rbanner}\}@habana.ai}}\\
\small{\texttt{{kfiryehud@gmail.com}}}\\
}

\begin{document}

\maketitle

\begin{abstract}
Quantization has established itself as the primary approach for decreasing the computational and storage expenses associated with Large Language Models (LLMs) inference. The majority of current research emphasizes quantizing weights and activations to enable low-bit general-matrix-multiply (GEMM) operations, with the remaining non-linear operations executed at higher precision. In our study, we discovered that following the application of these techniques, the primary bottleneck in LLMs inference lies in the softmax layer. The softmax operation comprises three phases: exponent calculation, accumulation, and normalization, Our work focuses on optimizing the first two phases. We propose an analytical approach to determine the optimal clipping value for the input to the softmax function, enabling sub-4-bit quantization for LLMs inference. This method accelerates the calculations of both $e^x$ and $\sum(e^x)$ with minimal to no accuracy degradation. For example, in LLaMA1-30B, we achieve baseline performance with 2-bit quantization on the well-known "Physical Interaction: Question Answering" (PIQA) dataset evaluation. This ultra-low bit quantization allows, for the first time, an acceleration of approximately 4x in the accumulation phase. The combination of accelerating both $e^x$ and $\sum(e^x)$ results in a 36.9\% acceleration in the softmax operation. A reference implementation\footnote{\url{https://github.com/Anonymous1252022/EXAQ}} is provided. 
\end{abstract}

\section{Introduction}
In recent years, the landscape of natural language processing (NLP) has been transformed by large language models (LLMs), showcasing unparalleled capabilities in contextual understanding and common sense reasoning. These capabilities are particularly evident as models are scaled up, driving research efforts towards further enlarging model dimensions \cite{gpt,llama2}. However, the substantial size of modern LLMs imposes considerable computational demands, making them resource-intensive in terms of training, fine-tuning, and inference processes. Consequently, there has been a surge in efforts to alleviate memory consumption and computational requirements. Among the promising approaches is quantization, a technique that involves representing parts of the model with lower bit widths, thereby reducing resource usage without compromising performance.

The foundation of LLMs lies in the attention mechanism \cite{attention}, which encompasses intensive general-matrix-multiply (GEMM) operations, coupled with non-linear operations like softmax. Consequently, prior quantization studies have primarily focused on reducing GEMM operations to 8 \cite{Peng2023FP8LMTF} or 4 \cite{Chmiel4bit, Frantar2022GPTQAP} bits with minimal to no degradation, showcasing significant advancements in this area. Additionally, modern hardware accelerators like Gaudi2 \cite{Habana} and H100 \cite{h100} support accelerated 8-bit (FP8) GEMMs, further emphasizing advancements in this area of quantization and diminishing the computational burden of GEMM operations, thus alleviating them from being the primary computational load. 

Once the bottleneck of GEMMs has been alleviated, attention has shifted toward reducing the computational demands of the softmax operation, which can account for more than 30\% of the total inference time. Efforts to accelerate softmax operations within the attention mechanism have predominantly revolved around quantizing the entire dynamic range of softmax inputs to 16 or 8 bits \cite{energy-eff-att-softmax-acc-for-qtrans}. However, this approach underscores the necessity for novel methods to enhance the efficiency of the softmax layer in terms of runtime, bandwidth, and memory usage, while maintaining accuracy. Our research indicates that current softmax acceleration via quantization is sub-optimal, suggesting that substantial performance improvements can be attained by tailoring the quantization optimization process to the specific properties of the softmax layer. 

The softmax operation comprises three main parts: (1) Exponent calculation: This involves taking the exponent of each input element. (2) Accumulation: The exponentiated values are then summed together. (3) Normalization: Each exponentiated value is divided by the sum to obtain the final softmax probabilities.
In this work, we are able to accelerate both steps (1) and (2) by quantizing, for the first time, the input to the exponent to below 4 bits.

First, we analyze the quantization error in the context of the exponential operation, comparing $e^X$ to $e^{X_q}$, where ${X_q}$ represents the quantized version of the input $X$. Subsequently, we introduce a pioneering approach to input quantization, coined "exponent-aware quantization" (EXAQ). This methodology presents an analytical model that strategically focuses on minimizing the quantization error after the exponent operation, directly targeting the exponential attributes of the softmax function. Lastly, we leverage the low-bit characteristics and propose a technique to unite the summation phase through a lookup table (LUT) operation, facilitating acceleration by approximately 4x. When we combine EXAQ with the accelerated accumulation we get an acceleration of 36.9 \% in the softmax operation.

\paragraph{Our paper introduces several key contributions:}

\begin{itemize}
\item We highlight the softmax layer as a significant computational bottleneck in modern neural networks. 

\item We propose an analytical approach to quantize the input to the exponent to below 4 bits, thereby enabling the utilization of a lookup table (LUT) based approach. This method notably diminishes the cycle consumption for computing $e^x$ to a single cycle. In contrast to FP32/BF16/FP16 formats, where creating a reasonably sized table is impractical.

\item We propose a technique to leverage the low-bit quantization of the softmax inputs and consolidate 4 consecutive summations into a lookup table (LUT), enabling up to 4x acceleration of the denominator accumulation process.

\item Our method achieves state-of-the-art accuracy for low-bit quantization of the softmax operation in LLMs. With 2-bit quantization, it reaches baseline accuracy in several tasks with no degradation, and when averaging across all tasks, it shows an average degradation of only $1.9\%$. This exceptional efficiency allows for the creation of an exceedingly compact LUT with just 4 entries, rendering our approach highly suitable for deployment on edge devices with extremely limited computational resources. 

\end{itemize}

\section{Motivation}
\label{sec:motivation}
This section aims to illustrate the considerable computational demand imposed by the softmax operation, highlighting the advantages of improving its runtime efficiency.
To establish a strong foundation for our argument, we conduct experiments to measure the runtime consumption using the "LLaMA-2-7B" LLM model on the Gaudi-2 accelerator, which is equipped with a high-speed network card for optimal performance. 

In Fig.\ref{fig:runtime_pie_chart}, we depict the proportion of time allocated to each operation during the model's execution in BF16 format. This graphical representation emphasizes the softmax layer as the main computational bottleneck. With GEMM operations functioning in BF16 format, the softmax layer consumes 39\% of the total runtime, while the GEMM operations contribute to 24\% of the runtime. Furthermore, given the advancement of modern accelerators supporting FP8 GEMMs acceleration, we anticipate that the softmax operation will consume an even larger portion of the total runtime.

To conclude, accelerating the softmax operation, particularly through techniques like quantization, has the potential to significantly increase the runtime efficiency of LLMs.


\section{Exponent-Aware Quantization (EXAQ)}
\label{sec:exaq}

In this section, we introduce a novel quantization method, "exponent-aware quantization" (EXAQ), specifically designed for softmax inputs. The softmax function, defined as $\text{softmax}(x_i) = \frac{e^{x_i}}{\sum_{j} e^{x_j}}$, operates by exponentiating its input logits and normalizing these values to form a probability distribution. Moreover, usually for numeric stability the maximum of x is subtracted before the exponent function. Our method focuses on minimizing the quantization error of the exponentiated outputs, ensuring a more precise representation of the softmax function's output. We manipulate the quantization in the input domain but target the mean squared error of the exponentiated outputs, minimizing $\text{MSE}(e^x, e^{Q(x)})$.

Inspired by the ACIQ paper \cite{DBLP:conf/nips/BannerNS19-aciq}, we limit the range of the tensor by clipping its values. While this introduces some distortion to the original tensor, it significantly reduces the rounding error in the part of the distribution containing most of the information. Since $x < 0$,  we set a threshold $C < 0$, so that if $x < C$, then $x = C$. Clipping is particularly useful because it preserves the less negative values, which after exponentiation become significantly larger compared to very negative values that become negligible after the exponential function is applied. Values in the range are quantized on a smaller scale, improving resolution for the more common and important values. The method approximates the optimal clipping value analytically from the distribution of the tensor by minimizing the MSE between $e^x$ and $e^{Q(x)}$. This analytical threshold is simple to use during run-time and can easily be integrated with other quantization techniques.

\subsection{Problem Formulation}
We begin by modifying the inputs for the function $e^x$ through the subtraction of the maximum value, $\max(x)$, from these inputs. Thus, it is assumed that $x \leq 0$. The MSE due to quantization and clipping can be expressed as a sum of two integrals: one for the quantization error for $x \in [C, 0]$ and another for the clipping error for $x < C$. The quantization error integral is given by:
\begin{equation}
\text{MSE}_{\text{quant}} = \int_C^0 (e^{\text{Q}(x)} - e^x)^2 \cdot f(x) \, dx,
\end{equation}
where $f(x)$ represents the probability density function of $x$, assumed to be gaussian distributed with mean $\mu$ and standard deviation $\sigma$. The clipping error integral is:
\begin{equation}
\text{MSE}_{\text{clip}} = \int_{-\infty}^C (e^C - e^x)^2 \cdot f(x) \, dx 
\end{equation}
Thus, total MSE is given by 
\begin{align}
\label{total_MSE}
\text{MSE} &=   \text{MSE}_{\text{clip}} + \text{MSE}_{\text{quant}}\\
&=   \int_{-\infty}^C (e^C  - e^{x})^2 \cdot f(x) \, dx + \int_C^0 (e^{\text{Q}(x)} - e^x)^2 \cdot f(x) \, dx.
\end{align}

\noindent
\begin{minipage}{0.48\textwidth}
\begin{figure}[H]
    \centering
    \includegraphics[width=0.8\linewidth]{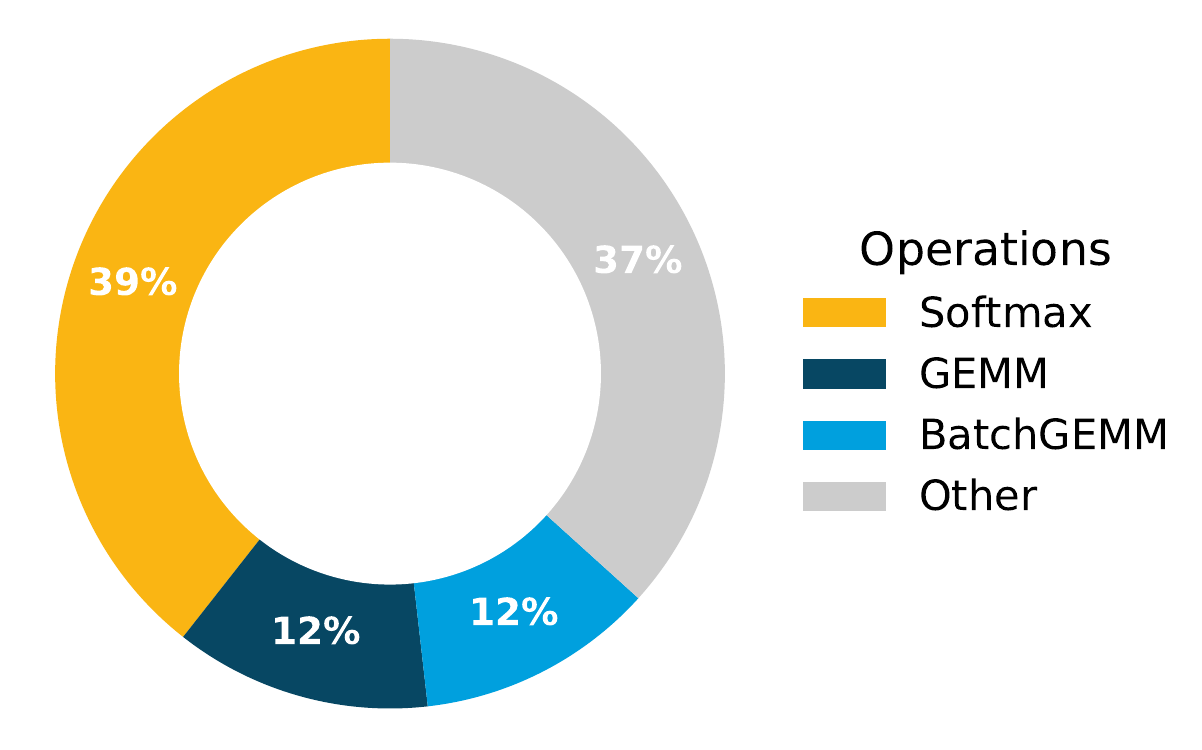}
    \caption{Distribution of runtime consumption by the layer type. The chart illustrates the proportional runtime spent on various layer types during model execution, highlighting the significant computational burden imposed by the softmax layer, which accounts for $39\%$ of the total runtime.}
    \label{fig:runtime_pie_chart}
\end{figure}
\end{minipage}\hfill
\begin{minipage}{0.48\textwidth}
\begin{figure}[H]
    \centering
    \includegraphics[width=\linewidth]{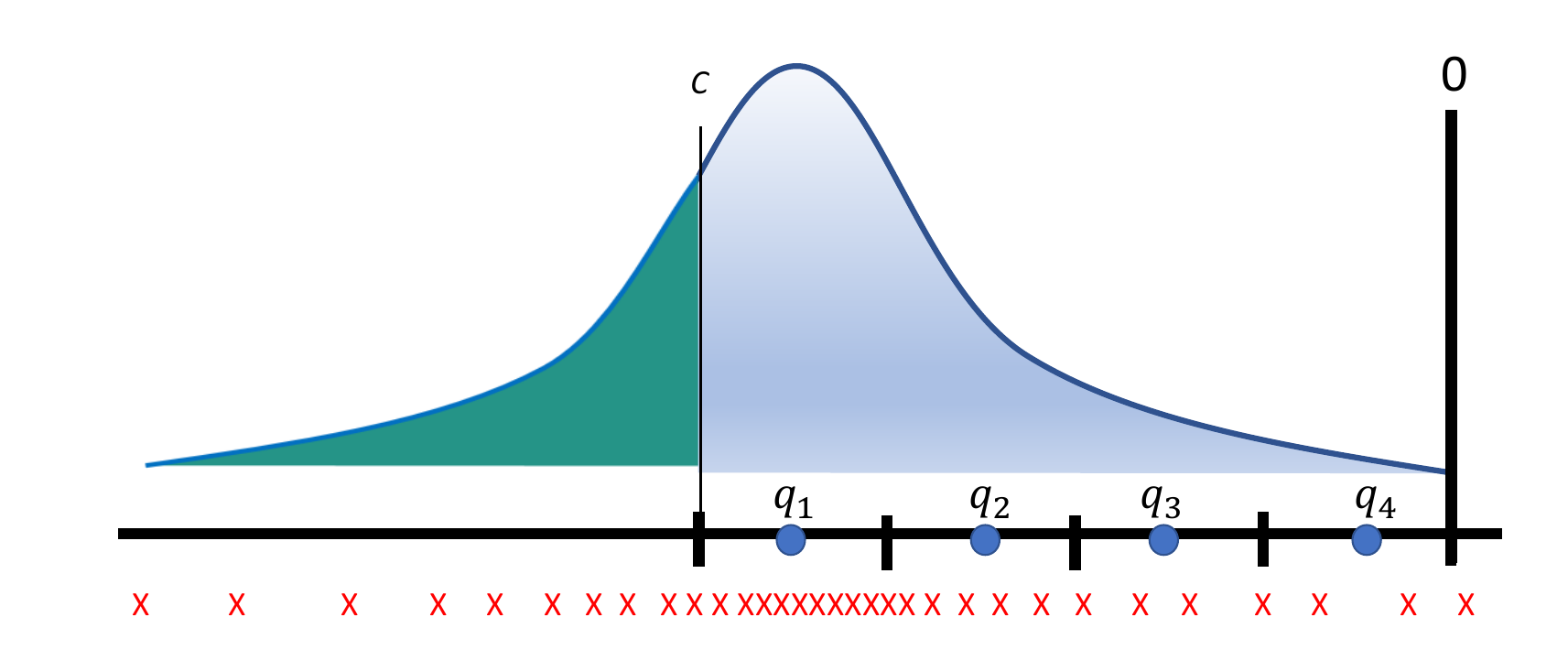}
    \vspace{0.55mm}
    \caption{Illustration of the distortion at the output of $e^x$ due to the quantization and clipping of the inputs. The clipping value $C$ is the threshold we aim to optimize. A very negative $C$ reduces clipping error but increases quantization error. The total mean squared error is the sum of these two contributions.}
    \label{fig:ilus_distortion}
\end{figure}
\end{minipage}


In Fig. \ref{fig:ilus_distortion} we present an illustration of the distortion of the proposed scheme. Before we get into the calculation of the mean squared error due to quantization, it is important to define the quantization process. We approximate the quantized value \(\text{Q}(x)\) as \(x + \epsilon\), where \(\epsilon\) represents the quantization noise. This noise is assumed to be drawn from a uniform distribution within the range \([- \Delta/2, \Delta/2]\), where \(\Delta\) is the quantization step size. For an \(M\)-bit integer quantization, the quantization step size \(\Delta\) is defined as \(\Delta = \frac{0-C}{2^M}\), accommodating the range of input values that need to be quantized. Given this quantization process, the MSE due to quantization can be analyzed as follows:

\begin{align}
\text{MSE}_{\text{quant}} &= \int_C^0 (e^{\text{Q}(x)} - e^x)^2 \cdot f(x) \, dx, \\
&= \frac{1}{\Delta} \int_{-\Delta/2}^{\Delta/2} \int_C^0 (e^{x + \epsilon} - e^x)^2 \cdot f(x) \, d\epsilon \, dx, \quad \\
&= \frac{1}{\Delta} \int_{-\Delta/2}^{\Delta/2} \int_C^0 (e^x + \epsilon e^x - e^x)^2 \cdot f(x) \, d\epsilon \, dx, \quad  (e^{x + \epsilon} \approx e^x + \epsilon e^x) \\
&= \frac{1}{\Delta} \int_{-\Delta/2}^{\Delta/2} \int_C^0 (\epsilon e^x)^2 \cdot f(x) \, d\epsilon \, dx, \\
&= \frac{1}{\Delta} \int_{-\Delta/2}^{\Delta/2} \epsilon^2 \, d\epsilon \int_C^0 (e^x)^2 \cdot f(x) \, dx \quad  \\
&= \frac{1}{\Delta} \left[ \frac{\epsilon^3}{3} \right]_{-\Delta/2}^{\Delta/2} \int_C^0 e^{2x} \cdot f(x) \, dx \\
\label{quant}
&= \frac{\Delta^2}{12} \int_C^0 e^{2x} \cdot f(x) \, dx \quad \end{align}

Substituting Equation \ref{quant} into \ref{total_MSE} we conclude that
\begin{align}
\label{integrals}
\text{MSE} &=  \text{MSE}_{\text{quant}} + \text{MSE}_{\text{clip}}\\
&= \frac{\Delta^2}{12} \int_C^0 e^{2x} \cdot f(x) \, dx  +  \int_{-\infty}^C (e^C - e^x)^2 \cdot f(x) \, dx \\
&= -\frac{C^2}{12\cdot 2^{2M}} \int_C^0 e^{2x} \cdot f(x) \, dx  + \int_{-\infty}^C (e^C - e^x)^2 \cdot f(x) \, dx 
\end{align}

\begin{figure}[htbp]
    \centering
    \begin{subfigure}[b]{0.49\textwidth}
        \includegraphics[width=\textwidth]{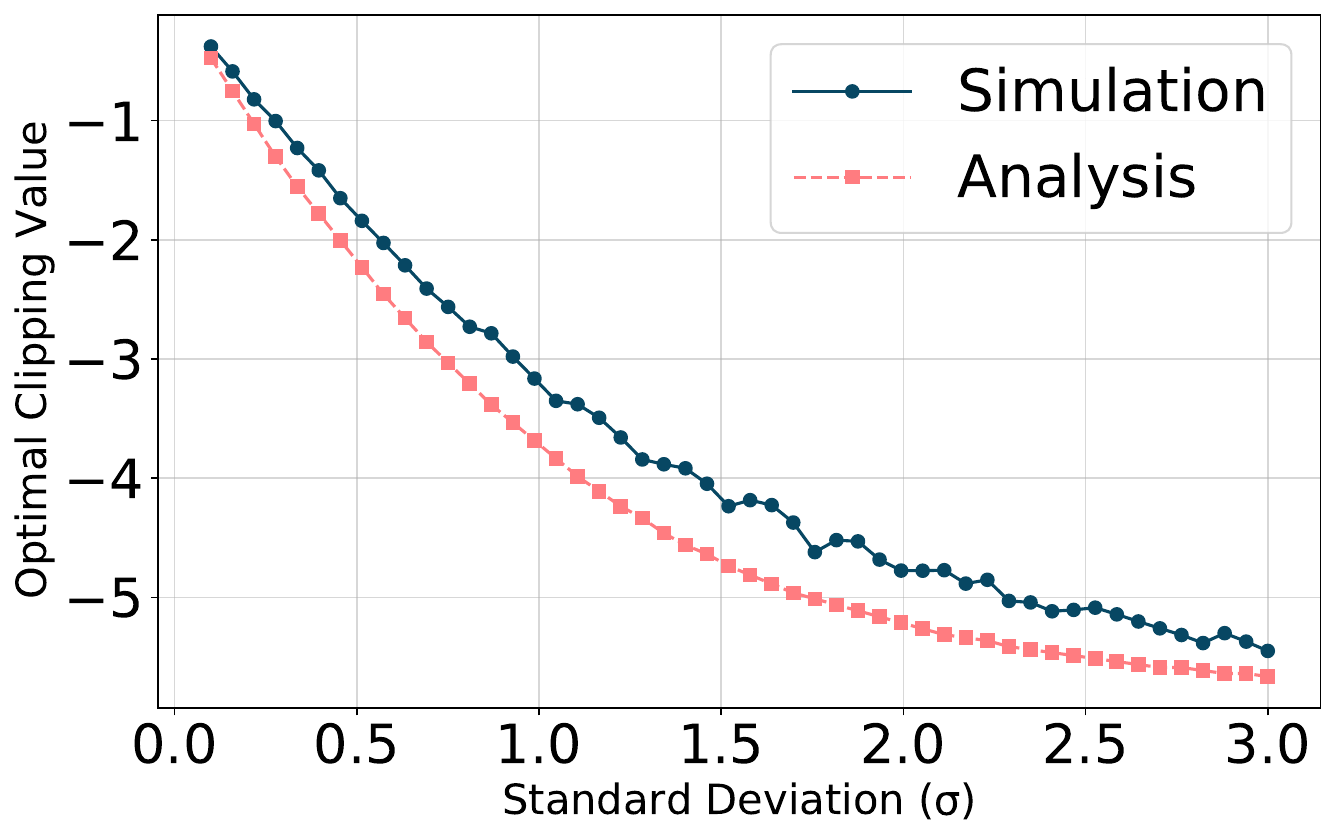}
        \caption{2-bit}
    \end{subfigure}
    \begin{subfigure}[b]{0.49\textwidth}
        \includegraphics[width=\textwidth]{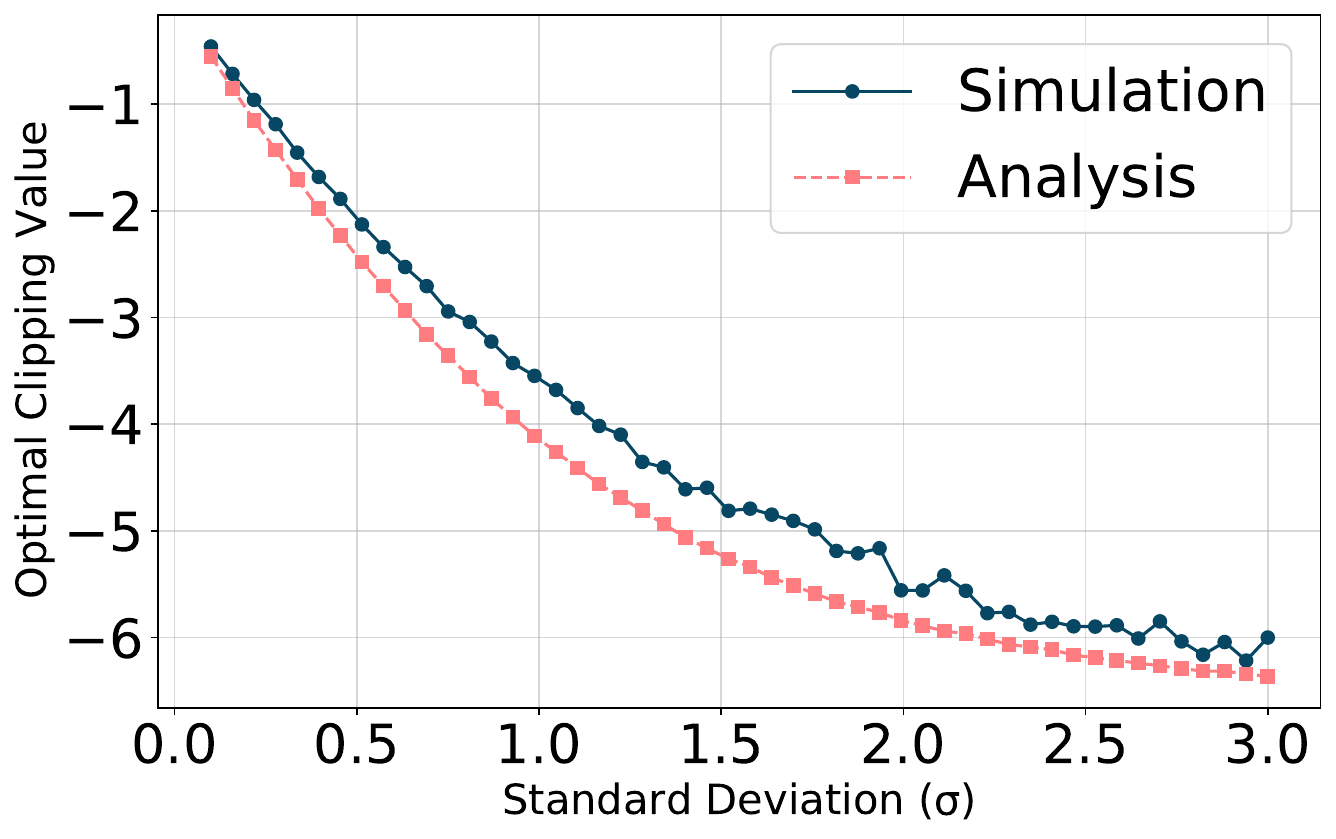}
        \caption{3-bit}
    \end{subfigure}
   \caption{Optimal clipping value vs. standard deviation of softmax input for different bit widths. The analysis and simulation results agree, demonstrating the accuracy of the analytical model. The simulation was conducted by generating 1000 samples from a normal distribution with mean 0 and various standard deviations.}
   \label{fig:clipping_values}
\end{figure}

Solving equation \ref{integrals} numerically for bit-widths $M = 2, 3$ and finding the optimal clipping value that minimizes MSE yields optimal clipping values as functions of the standard deviation ($\sigma$). These results are visualized in Figure \ref{fig:clipping_values} for a normal density function with standard deviation $\sigma$.

Finally, we use a linear approximation to estimate the optimal clipping values in the range [0.9, 3.4], where most standard deviations occur in practice (as seen in Figure \ref{fig:standard_deviation_histogram}). This approach allows us to avoid maintaining a detailed table that maps the standard deviation to optimal clipping values ($C^{*}$). Instead, we focus on keeping only two variables (slope and intercept) to estimate the linear approximation for the optimal clipping value. This way, once we have the standard deviation, we can immediately calculate the estimated optimal clipping value using the following table:


\begin{table}[h]
\caption{Linear approximation for optimal clipping value ($C^{*}$)}
  \label{table:linear_appro}
  \centering
  \begin{tabular}{cc}  
    \toprule
   Number of bits (M) & $C^{*}$ \\  
    \midrule
    2  & $-1.66 \cdot \sigma -1.85$ \\
    3  & $-1.75 \cdot \sigma -2.06$ \\  
    \bottomrule
  \end{tabular}
  \vspace{-0.4cm}
\end{table}

\section{Algorithm implementation}
\label{sec:algo}

The computation of the softmax function is typically broken down into three essential steps: (1)  Exponent calculation: this step involves computing $e^x$ for each input $x$ (2) Accumulation: this step involves summing all the exponential values to form the denominator of the softmax function, and (3) Normalization: this step divides each exponential value by the computed sum to produce the final softmax output. Our algorithm primarily focuses on steps (1) and (2), leveraging the ultra-low precision of softmax inputs facilitated by the EXAQ method, to accelerate these two operations.
Fig.\ref{fig:two_softmax_algos} compares the original softmax algorithm and our optimized 2-bit version, highlighting the computational efficiencies achieved.

\subsection{Exponent calculation} 

We replace the traditional direct exponent calculation (line 4 in Algo.\ref{alg:two_bit_softmax}) with the following two steps: (1) We quantize each element in the normalized tensor into a 2-bit integer (line 4 in Algo.\ref{alg:two_bit_softmax}). (2) We utilize pre-computed values from a lookup table $LUT_{exp}$ to derive the exponents of the quantized values (lines 5-6 in Algo.\ref{alg:two_bit_softmax}). This $LUT_{exp}$ maps between all possible quantized values and their resulting exponents and is notably compact as it needs to store only 4 values. This approach not only reduces memory usage but also speeds up the algorithm since the exponential values can be retrieved in a single cycle.\footnote{While direct exponent calculation typically takes 5-12 cycles, depending on the hardware design.}


\subsection{Accelerated denominator accumulation}

\begin{figure}[h]
\centering
\includegraphics[width=0.95\linewidth]{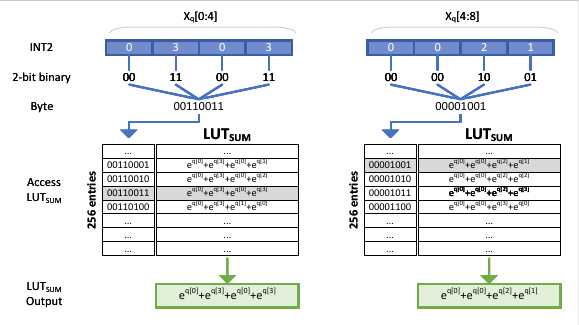}
\caption{ Illustration of the proposed accelerated denominator accumulation.
The $LUT_{sum}$ lookup table contains pre-computed values of sums of the exponents of 4 consecutive quantized tensor elements.
In the left example, the integer representations of the quantized values are $X_q[0:4]=[0,3,0,3]$, and their corresponding floating-point representations are $[q[0], q[3], q[0], q[3]]$. The lookup key is constructed by concatenating the 2-bit counterparts of the 4 integer representations into a single byte.
}
\vspace{-0.5cm}
    \label{fig:lut_scheme}
\end{figure}

Originally, the accumulation process within the softmax layer’s denominator requires summing up $N$ exponential outputs (Algo.\ref{alg:original_softmax} lines 7-12). In contrast, our algorithm requires only $N/4$. Since the input tensor’s values are quantized to 2 bits, each byte can now represent 4 values. We utilize an additional lookup table, $LUT_{sum}$, that maps between all possible combinations of 4 quantized values to the value of the sum of their exponents. 
We calculate the denominator using the following steps. First, we divide the quantized tensor $x_q$ into $N/4$ sequences ($s$ denotes a sequence) of 4 values $[s_0, s_1, … s_{N/4-1}]=[x_q[0:4], x_q[4:8], … x_q[N-4:N]]$. Next, we apply $LUT_{sum}$ on each $s_i$ to obtain the sum of the exponents of the corresponding sequence. Finally, we sum the resulting $N/4$ values. In Figure \ref{fig:lut_scheme} we show an illustration of the proposed accelerated denominator accumulation.



Our algorithm simplifies the accumulation step from 4 separate accumulations (4 cycles) to a single LUT access (1 cycle), enhancing the speed of the denominator calculation by a factor of 4.
A pseudo-code of the proposed algorithm appears in Algo.\ref{alg:two_bit_softmax} (lines 10:13). The entire denominator accumulation process is completed within $N/4$ iterations, compared to the original algorithm shown in Algo.\ref{alg:original_softmax} (lines 9:12), which requires $N$ iterations for the same purpose.
This algorithm also decouples the exponential computation from the denominator accumulation, allowing these steps to be executed concurrently, as opposed to the original softmax algorithm. 
Moreover, this approach can be extended to a 4-bit quantization, providing a 2x acceleration, as each byte can accommodate two 4-bit values.

\begin{figure}[t]
    \centering
\begin{minipage}{0.45\textwidth}
        \begin{algorithm}[H]
            \caption{Original softmax algorithm}
            \label{alg:original_softmax}  
            \begin{algorithmic}[1]
            \State \textbf{Input:} input tensor $x$
            \newline
            \State \textbf{Output:} softmax tensor $out(x)$
            \State Normalize input tensor: $x = x - \max(x)$
            \newline 
            \newline
            \For{$i = 1$ to $\text{size}(x)$}
            \State \textcolor{red}{$e[i] = e^{x[i]}$} \textcolor{red}{\Comment{{multi cycle op}}}
            \EndFor
            \newline  
            Denominator accumulation:
            \State $sum = 0$
            \State$i=1$
            \While {$i\leq \text{size}(x)$}
            
            \State $sum = sum + e[i]$
            \State \textcolor{red}{$i+=1$}
            \EndWhile
            \For{$i = 1$ to $\text{size}(x)$}
                \State $out(x[i])= e[i] / sum$
            \EndFor
            \end{algorithmic}
            \end{algorithm}
    \end{minipage}
    \hfill
    \begin{minipage}{0.45\textwidth}
        \begin{algorithm}[H]
            \caption{2-bit softmax algorithm}
            \label{alg:two_bit_softmax}  
            \begin{algorithmic}[1]
            \State \textbf{Input:} input tensor $x$, $LUT_{\text{exp}}$, $LUT_{\text{sum}}$, scale, offset, clip
            \State \textbf{Output:} softmax tensor $out(x)$
            \State Normalize input tensor: $x = x - \max(x)$
            \State quantize $x$: \textcolor{CC_GREEN}{\Comment{3 cycles op}} \newline 
            $x_q = \text{Q}(x, \text{scale},\text{offset}, \text{clip})$
            \For{$i = 1$ to $\text{size}(x)$}
            \State \textcolor{CC_GREEN}{$e[i] = LUT_{\text{exp}}[x_{q}[i]]$} \textcolor{CC_GREEN}{\Comment{1 cycle op}}
            \EndFor
            \newline 
            Denominator accumulation:
            \State $sum = 0$
            \State$i=1$
            \While{$i \leq \text{size}(x)$}
            \State $sum = sum +             \textcolor{CC_GREEN}{LUT_{\text{sum}}[x_{q}[i:i+3]]}$
            \State \textcolor{CC_GREEN}{$i += 4$} 
            \EndWhile
            \For{$i = 1$ to $\text{size}(x)$}
                \State $out(x[i])= e[i] / sum$
            \EndFor
            \end{algorithmic}
            \end{algorithm}
    \end{minipage}
    \hfill
    \caption{Comparison of softmax algorithms: Algorithm \ref{alg:original_softmax} details the original softmax computation method, involving multiple cycle exponential operations and $N$ accumulations in the denominator. Algorithm \ref{alg:two_bit_softmax} introduces a 2-bit optimized version using lookup tables (LUTs), which involves a single cycle exponential operation and $N/4$ accumulations in the denominator.}
    \label{fig:two_softmax_algos}
\end{figure}

\section{Experiments}
\label{sec:experiments}

This section details the experimental framework used to evaluate the performance of our quantization method. We evaluate the accuracy across various language tasks, comparing our softmax quantization method (EXAQ) to the naive quantization method (NAIVE). Our method achieves state-of-the-art accuracy scores in almost all experiments.

\subsection{Accuracy experiments}

\subsubsection{Experimental settings}
\label{sec:exp_setting}

Our accuracy experiments focus on the inference setting and are conducted on 8 RTX A6000 GPUs, utilizing a batch size of 4 for all evaluations. We use the LLaMA-1 models \cite{touvron2023llama}, specifically the 7B, 13B, 30B and 70B variants, and assess these models on a variety of question-answering and reasoning tasks, such as BoolQ \cite{boolq_ref} and WinoGrande \cite{winogrande_ref}.  
 The experiments are implemented using modifications to the lm-evaluation-harness \cite{eval-harness}, an open-source framework that utilizes pre-trained models from the \emph{HuggingFace Project} \footnote{\url{https://huggingface.co/docs/transformers/main/en/model_doc/llama}}.

\paragraph{Quantization settings.}
The softmax input quantization function parameters need to be tuned based on tensor statistics collected from a calibration set. In our experiments, we run a calibration set of size 100 by running 25 iterations each with a batch size of 4.

\subsubsection{Inference accuracy evaluation}

Table \ref{table:inference_accuracy_llama1} provides an insightful visual comparison of inference accuracy using different scales of LLaMA models (7B, 13B, 30B and 70B parameters) across 7 NLP tasks: BoolQ \cite{boolq_ref}, HellaSwag \cite{hellaswag_ref}, PIQA \cite{piqa_ref}, WinoGrande \cite{winogrande_ref}, ARC Challenge \cite{Clark2018ThinkYH_arc}, ARC Easy \cite{Clark2018ThinkYH_arc} and OpenBookQA \cite{Mihaylov2018CanAS_openbook}. All models have their softmax inputs quantized to 2-bit and 3-bit precision using our method EXAQ and the NAIVE method. EXAQ calculates the optimal clipping parameter using the standard deviation ($\sigma$) of the input tensor, as detailed in Table.\ref{table:linear_appro}, while NAIVE sets the clipping parameters by averaging the tensor's minimum and maximum values.
Our method achieves state-of-the-art accuracy scores in almost all experiments (noted with bold marks in Table.\ref{table:inference_accuracy_llama1}).
With 3-bit softmax inputs, EXAQ reaches the baseline within $0.65\%$ on average, with $43\%$ of the results either meeting or exceeding the baseline accuracy (noted with green color in Table.\ref{table:inference_accuracy_llama1}). With 2-bit softmax inputs, EXAQ approaches the baseline within $1.9\%$ on average and reaches the baseline without degradation in several tasks. Additional experiments appear in section \ref{secApp:exp}.

\begin{table}[h!]
\caption{Inference accuracy evaluation for different LLaMA-1 models across various tasks.}
\centering
\small

\begin{subtable}{\textwidth}
\centering
\scalebox{0.9}{
\setlength\tabcolsep{3pt}
\begin{tabular}{p{1.5cm}|p{0.65cm}|S[table-format=2.1]*{6}{S[table-format=2.1]}|>{\columncolor{lightgray1}}S[table-format=2.1]}
\toprule
\textbf{Q method} & \textbf{Prec.} & \textbf{BoolQ} & \textbf{HellaSwag} & \textbf{PIQA} & \textbf{WinoGrande} & \textbf{ARC Challenge} & \textbf{ARC Easy} & \textbf{OpenBookQA} & \textbf{avg score} \\
\midrule
NONE               & \centering BF16 & 75.1 & 76.2 & 79.2 & 69.6 & 45.1 & 73.2 & 44.4 & 66.1 \\
\cmidrule{1-10}
NAIVE            & \centering \multirow{2}{*}{INT2} & 46.6 & 54.5 & 69.0 & 55.6 & 30.5 & 55.9 & 36.8 & 49.8 \\
\textbf{EXAQ}       &  & \textbf{73.0} & \textbf{72.9} & \textcolor{CC_GREEN}{\textbf{79.2}} & \textcolor{CC_GREEN}{\textbf{69.6}} & \textbf{43.9} & \textbf{72.4} & \textbf{41.4} & \textbf{64.6} \\
\cmidrule{1-10}
NAIVE            & \centering \multirow{2}{*}{INT3} & 71.3 & 73.7 & 78.5 & 67.2 & 42.8 & 71.0 & 43.4 & 63.9 \\
\textbf{EXAQ}       &  & \textcolor{CC_GREEN}{\textbf{75.1}} & \textbf{74.8} & \textcolor{CC_GREEN}{\textbf{79.3}} & \textcolor{CC_GREEN}{\textbf{69.7}} & \textbf{44.2} & \textbf{72.9} & \textbf{43.8} & \textbf{65.7} \\
\bottomrule
\end{tabular}
}
\caption{LLaMA-1-7B}
\label{subtab:llama1_7b}
\end{subtable}


\begin{subtable}{\textwidth}
\centering
\scalebox{0.9}{
\setlength\tabcolsep{3pt}
\begin{tabular}{p{1.5cm}|p{0.65cm}|S[table-format=2.1]*{6}{S[table-format=2.1]}|>{\columncolor{lightgray1}}S[table-format=2.1]}
\toprule
\textbf{Q method} & \textbf{Prec.} & \textbf{BoolQ} & \textbf{HellaSwag} & \textbf{PIQA} & \textbf{WinoGrande} & \textbf{ARC Challenge} & \textbf{ARC Easy} & \textbf{OpenBookQA} & \textbf{avg score} \\
\midrule
NONE               & \centering BF16 & 77.7 & 79.1 & 80.1 & 72.8 & 47.9 & 74.8 & 44.6 & 68.2 \\
\cmidrule{1-10}
NAIVE            & \centering \multirow{2}{*}{INT2} & 56.1 & 48.5 & 62.3 & 55.6 & 26.6 & 51.7 & 38.4 & 48.5 \\
\textbf{EXAQ}       &  & \textbf{73.9} & \textbf{75.8} & \textbf{79.4} & \textbf{71.7} & \textcolor{CC_GREEN}{\textbf{47.9}} & \textbf{73.7} & \textcolor{CC_GREEN}{\textbf{44.6}} & \textbf{66.7} \\
\cmidrule{1-10}
NAIVE            & \centering \multirow{2}{*}{INT3} & 73.7 & 77.3 & 79.1 & \textbf{71.8} & 45.1 & 72.6 & 44.6 & 66.3 \\
\textbf{EXAQ}       &  & \textbf{76.1} & \textbf{78.0} & \textcolor{CC_GREEN}{\textbf{80.3}} & 71.7 & \textcolor{CC_GREEN}{\textbf{47.9}} & \textbf{74.6} & \textcolor{CC_GREEN}{\textbf{45.8}} & \textbf{67.8} \\
\bottomrule
\end{tabular}
}
\caption{LLaMA-1-13B}
\label{subtab:llama1_13b}
\end{subtable}


\begin{subtable}{\textwidth}
\centering
\scalebox{0.9}{
\setlength\tabcolsep{3pt}
\begin{tabular}{p{1.5cm}|p{0.65cm}|S[table-format=2.1]*{6}{S[table-format=2.1]}|>{\columncolor{lightgray1}}S[table-format=2.1]}
\toprule
\textbf{Q method} & \textbf{Prec.} & \textbf{BoolQ} & \textbf{HellaSwag} & \textbf{PIQA} & \textbf{WinoGrande} & \textbf{ARC Challenge} & \textbf{ARC Easy} & \textbf{OpenBookQA} & \textbf{avg score} \\
\midrule
NONE               & \centering BF16 & 82.8 & 82.6 & 81.2 & 75.3 & 53.0 & 78.9 & 48.4 & 71.9 \\
\cmidrule{1-10}
NAIVE            & \centering \multirow{2}{*}{INT2} & 54.0 & 64.9 & 72.1 & 60.0 & 33.8 & 64.4 & 42.6 & 56.0 \\
\textbf{EXAQ}       &  & \textbf{80.6} & \textbf{78.1} & \textcolor{CC_GREEN}{\textbf{81.3}} & \textbf{74.3} & \textbf{51.9} & \textbf{77.9} & \textbf{47.8} & \textbf{70.3} \\
\cmidrule{1-10}
NAIVE            & \centering \multirow{2}{*}{INT3} & \textbf{81.0} & 81.4 & 81.2 & 75.1 & 51.8 & 78.5 & 47.0 & 70.9 \\
\textbf{EXAQ}       &  & 80.6 & \textbf{80.7} & \textcolor{CC_GREEN}{\textbf{82.2}} & \textcolor{CC_GREEN}{\textbf{75.3}} & \textcolor{CC_GREEN}{\textbf{54.0}} & \textbf{78.8} & \textbf{48.0} & \textbf{71.4} \\
\bottomrule
\end{tabular}
}
\caption{LLaMA-1-30B}
\label{subtab:llama1_30b}
\end{subtable}


\begin{subtable}{\textwidth}
\centering
\scalebox{0.9}{
\setlength\tabcolsep{3pt}
\begin{tabular}{p{1.5cm}|p{0.65cm}|S[table-format=2.1]*{6}{S[table-format=2.1]}|>{\columncolor{lightgray1}}S[table-format=2.1]}
\toprule
\textbf{Q method} & \textbf{Prec.} & \textbf{BoolQ} & \textbf{HellaSwag} & \textbf{PIQA} & \textbf{WinoGrande} & \textbf{ARC Challenge} & \textbf{ARC Easy} & \textbf{OpenBookQA} & \textbf{avg score} \\
\midrule
NONE               & \centering BF16 & 84.9 & 84.2 & 82.3 & 77.2 & 55.5 & 79.9 & 47.4 & 73.0 \\
\cmidrule{1-10}
NAIVE            & \centering \multirow{2}{*}{INT2} & 39.2 & 61.7 & 68.6 & 55.0 & 32.9 & 64.2 & 39.8 & 51.6 \\
\textbf{EXAQ}       &  & \textbf{72.3} & \textbf{79.4} & \textbf{81.6} & \textbf{76.7} & \textbf{53.9} & \textbf{78.9} & \textbf{47.2} & \textbf{70.0} \\
\cmidrule{1-10}
NAIVE            & \centering \multirow{2}{*}{INT3} & \textbf{82.5} & 82.5 & 81.5 & 76.2 & 53.8 & 79.2 & 46.6 & \textbf{71.8} \\
\textbf{EXAQ}       &  & 77.7 & \textbf{82.7} & \textcolor{CC_GREEN}{\textbf{82.5}} & \textbf{77.0} & \textbf{54.7} & \textcolor{CC_GREEN}{\textbf{79.9}} & \textcolor{CC_GREEN}{\textbf{47.6}} & 71.7 \\
\bottomrule
\end{tabular}
}
\caption{LLaMA-1-65B}
\label{subtab:llama1_65b}
\end{subtable}
\label{table:inference_accuracy_llama1}
\vspace{-0.5cm}
\end{table}

\subsection{Runtime experiments}

We conducted runtime experiments to evaluate the overall performance of our algorithm, isolating the softmax operation to measure its runtime. Results are shown in Table.\ref{table:softmax_runtime}.
Our optimized algorithm demonstrates a significant improvement in runtime performance for the softmax operation, achieving an enhancement of $36.9\%$.

\begin{table}[h]
\caption{Softmax layer runtime performance}
  \label{table:softmax_runtime}
  \centering
  \begin{tabular}{lc}  
    \toprule
   Implementation & Average Runtime (ms) \\  
    \midrule
    Original algorithm (Algo.\ref{alg:original_softmax})  & 3.274 \\
    \textbf{Our algorithm (Algo.\ref{alg:two_bit_softmax}})  & 2.066 \\  
    \bottomrule
  \end{tabular}
\end{table}

\section{Related work}
\label{related_work}

\subsection{Language models acceleration} 

In the quest to optimize neural networks for practical deployment, particularly LLMs, studies like \cite{ibert, q8bert-zafrir_2019} have introduced innovative approaches to reduce computational demands. \cite{ibert}, for example, implements an integer-only quantization scheme for Transformers that conducts all inference operations with integer arithmetic, using INT32 for softmax inputs. Similarly, \cite{q8bert-zafrir_2019} applies selective quantization to Transformers, focusing specifically on GEMM layers while keeping softmax in FP32 precision. 

\subsection{Softmax acceleration} 

Other works focus on softmax acceleration as it has become a bottleneck in recent years for LLMs. \cite{basic-split-Sun2018AHS} proposes to use basic-split calculation method, which allows to split the exponentiation calculation of the softmax into several specific basics which are implemented by LUTs and multipliers. 
\cite{Li2018EfficientFPGA, Geng2018HardwareAwareSoftmaxAppr, Stevens2021SoftermaxHC, zhu2020precision-adjustable} compute the exponential operations of integer and fractional parts separately using a combination of LUTs and piecewise linear (PWL) function fitting. \cite{Li2018EfficientFPGA, Geng2018HardwareAwareSoftmaxAppr} also accelerate the division operation by replacing the divider with shifter units. All works use LUT much bigger than ours method (4 entries) except \cite{Geng2018HardwareAwareSoftmaxAppr} which requires a fine-tuning phase to achieve a small LUT size and is not applicable for the post-training quantization paradigm we work with. \cite{energy-eff-att-softmax-acc-for-qtrans} utilizes quantization-aware training to clip the range of the softmax inputs, subsequently quantizing them to INT8 and replacing the exponential operation with shifter units. \cite{Stevens2021SoftermaxHC} targets the softmax acceleration but requires a fine-tuning phase to mitigate major accuracy losses. \cite{DBLP:conf/iscas/GaoLL20-taylor} splits the exponential operation to high-bits/low-bits and uses Taylor series expansion to approximate the exponential calculation and reduce computation. \cite{Dao2022FlashAttentionFA} accelerate the attention mechanism by reducing the number of writes and reads operations between the HBM and SRAM.

The most closely related works to ours are \cite{DBLP:conf/hpca/HamJKOPSPLPLJ20-A3-low-high-bits} and \cite{vasyltsov2021-2D-LUT}, both of which aim to accelerate the softmax operation and, like our method, do not require a fine-tuning phase. 
\cite{DBLP:conf/hpca/HamJKOPSPLPLJ20-A3-low-high-bits} addresses only the exponent calculation acceleration, disregarding the denominator. In contrast, one of the key advantages of our approach is its significant improvement in the denominator accumulation, reducing the number of required accumulations by a factor of 4. Additionally, for the exponent calculation, \cite{DBLP:conf/hpca/HamJKOPSPLPLJ20-A3-low-high-bits} uses two 256-entry LUTs and a multiplication, taking 3 cycles, whereas our approach uses a single 4-entry LUT, requiring just one cycle. This significantly enhances both runtime and memory efficiency. A detailed comparison is in \ref{competitive_comparison_1}.
\cite{vasyltsov2021-2D-LUT} introduces two methods for softmax acceleration: one using two 1D-LUTs combined with a multiplier, and another using a combination of 1D-LUT and 2D-LUT, without a multiplier. Both methods, like ours, accelerate the exponent calculation by accessing one LUT in one cycle. However, our approach significantly accelerates the denominator accumulation by a factor of 4, while their method focuses on enhancing the division phase. The distinct enhancements of each method suggest a potential synergy, where our improved denominator accumulation could complement their division optimizations for a complete softmax enhancement. A detailed comparison is in \ref{competitive_comparison_2}



\section{Discussion}
\label{section:discussion}

\subsection{Summary} 
In this study, we analyze the execution time of various operations during LLMs inference and demonstrate that the softmax operation emerges as one of the primary bottlenecks. Moreover, we anticipate that as GEMM acceleration advances, the bottleneck will become even more critical.

Based on this conclusion, we introduce EXAQ - an analytical approach aimed at reducing the dynamic range of the input of the exponent, thereby enabling quantization below 4 bits and accelerating the exponent calculation. Additionally, leveraging ultra-low quantization, we propose a method to accelerate the accumulation step by up to 4 times.
The proposed full solution is able to get 36.9 \% acceleration in the softmax operation. 

We demonstrate that our proposed method achieves minimal to no degradation for the first time, in 2-bit and 3-bit quantization across various LLM sizes (7B, 13B, 30B and 70B) and a range of evaluated tasks.

\subsection{Limitations}
\label{subsec:limitations}
In our work, we focused on minimizing the quantization error of the exponential output. A more precise approach, however, would involve minimizing the quantization error of the softmax outputs or the entire attention block.  This alternative approach was not explored in the current research and is identified as an important avenue for future work. Additionally, our methodology was tested only during the inference stage of the model's lifecycle. Exploring its effects during the training phase remains an area for future investigation.


\bibliography{neurips_2024}
\bibliographystyle{plainnat}

.
\medskip

{
\small


\appendix
\newpage

\part*{Appendix}

\section{Further discussion}

\subsection{Broader impacts} 
\label{subsection:broader_impacts}
The acceleration of large language model (LLM) runtime significantly impacts modern life, particularly as tools like ChatGPT and Gemini become more integrated into daily use. Speeding up these models is essential for ongoing development and growth in this area, as it tackles a critical bottleneck: the processing speed and efficiency of the algorithms. Moreover, enhancing these models' speed and reducing their memory footprint not only improves their performance but also makes them more accessible to a wider audience. This allows more users to customize and advance these models for their specific needs and developments.

\section{Standard Deviation Range}
\begin{figure}[htbp]
    \centering
    \includegraphics[width=\textwidth
    ]{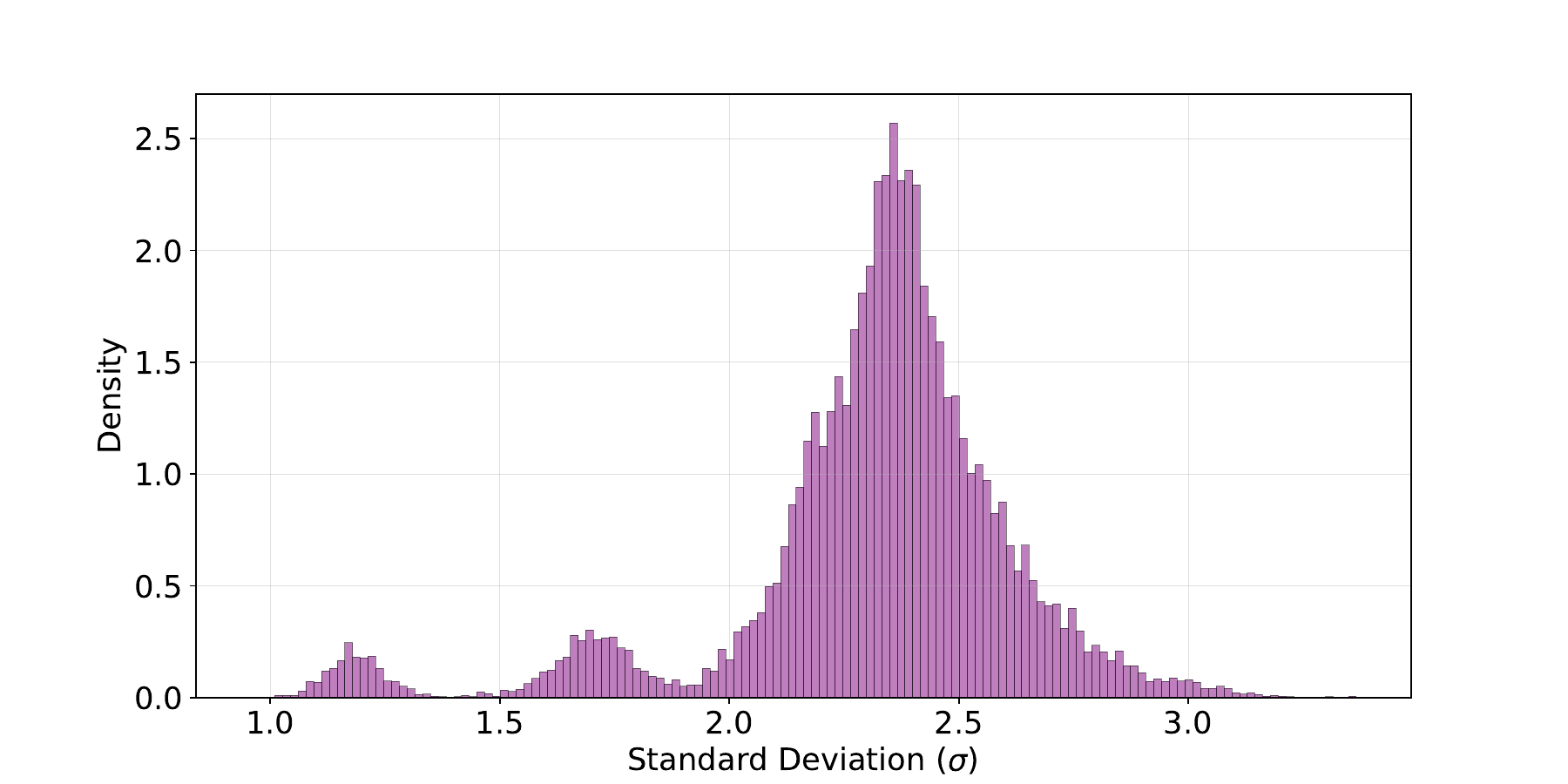}
   \caption{Standard deviation of softmax input collected across different layers and iterations.}
   \label{fig:standard_deviation_histogram}
\end{figure}

\section{Comparing our algorithm to the latest algorithms}

\subsection{A competitive comparison against \cite{DBLP:conf/hpca/HamJKOPSPLPLJ20-A3-low-high-bits}}
\label{competitive_comparison_1}

A key advantage of our approach is its significant improvement in denominator accumulation, reducing the number of required accumulations by a factor of 4, a notable acceleration, as the method proposed in \cite{DBLP:conf/hpca/HamJKOPSPLPLJ20-A3-low-high-bits} does not address this aspect.
The method in \cite{DBLP:conf/hpca/HamJKOPSPLPLJ20-A3-low-high-bits} quantizes FP16 inputs to 16-bit fixed-point and calculates exponents using two separate 256-entry LUTs, followed by a multiplication. 
This process requires 3 cycles for the two LUT accesses and the subsequent multiplication.
In contrast, our algorithm quantizes inputs to 2-bit integers and uses a single ultra-small LUT with only 4 entries. 
This streamlined approach reduces the process to just one cycle for the single LUT access, significantly enhancing both runtime and memory efficiency compared to the method in \cite{DBLP:conf/hpca/HamJKOPSPLPLJ20-A3-low-high-bits}.

\subsection{A competitive comparison against \cite{vasyltsov2021-2D-LUT}}
\label{competitive_comparison_2}

This work supports softmax inputs in integer format and introduces two methods to accelerate the softmax operation via approximation. The first method employs two 1-dimensional lookup tables (1D-LUTs) to approximate $e^x$ and $\frac{1}{x}$, combining these outputs with a multiplier to produce the final result. The second method combines a 1D-LUT and a 2D-LUT (2-dimensional lookup table).
In this approach, the output from the 1D-LUT and the results from the accumulated denominator are used as the indices [i, j] for the 2D-LUT, which directly contains the final softmax result, thereby eliminating the need for multiplication or division. 
However, this approach has been noted to cause an additional drop in accuracy. Additionally, a de-quantization phase is conducted if the next layer requires an FP format. To conclude, assuming the softmax inputs are in floating-point format, both our method and that of \cite{vasyltsov2021-2D-LUT} require an initial quantization phase.
Each approach utilizes LUTs to approximate $e^x$, with each requiring just one cycle for LUT access.
However, our work significantly accelerates the denominator accumulation phase by a factor of 4.
In contrast, \cite{vasyltsov2021-2D-LUT} enhances the normalization phase efficiency by combining a 1D-LUT with a multiplier or a direct use of a 2D-LUT.
The distinct enhancements made by each method suggest a potential synergy if integrated. 
Our improvements in denominator accumulation could complement the division optimizations made by \cite{vasyltsov2021-2D-LUT}, offering a complete enhancement to the softmax function.
Additionally, while \cite{vasyltsov2021-2D-LUT}'s process concludes with a de-quantization phase that requires additional computational steps, our method eliminates the need for this phase, reducing overall cycles, thereby providing an advantage to our approach.

\section{Additional Experiments}
\label{secApp:exp}
\subsection{Statistical significance measurements of results in Table \ref{table:inference_accuracy_llama1} (inference accuracy evaluation for LLaMA-1 models)}

\begin{table}[h!]
\caption{The standard deviation $(\sigma)$ over multiple runs of LLaMA-1. The mean values of these runs are presented in Table \ref{table:inference_accuracy_llama1} in the main paper.}
\centering
\small

\begin{subtable}{\textwidth}
\centering
\scalebox{0.9}{
\setlength\tabcolsep{3pt}
\begin{tabular}{p{1.5cm}|p{0.65cm}|S[table-format=2.1]*{6}{S[table-format=2.1]}}
\toprule
\textbf{Q method} & \textbf{Prec.} & \textbf{BoolQ} & \textbf{HellaSwag} & \textbf{PIQA} & \textbf{WinoGrande} & \textbf{ARC Challenge} & \textbf{ARC Easy} & \textbf{OpenBookQA} \\
\midrule
NONE             & FP16 & 0.76 & 0.43 & 0.95 & 1.29 & 1.45 & 0.91 & 2.22 \\
\cmidrule{1-9}
NAIVE            & \centering \multirow{2}{*}{INT2} & 0.87 & 0.50 & 1.08 & 1.40 & 1.34 & 1.02 & 2.16 \\
EXAQ     &  & 0.78 & 0.44 & 0.95 & 1.29 & 1.45 & 0.92 & 2.20 \\
\cmidrule{1-9}
NAIVE            & \centering \multirow{2}{*}{INT3} & 0.79 & 0.44 & 0.96 & 1.32 & 1.45 & 0.93 & 2.22 \\
EXAQ     &  & 0.76 & 0.43 & 0.95 & 1.29 & 1.45 & 0.91 & 2.22 \\
\bottomrule
\end{tabular}
}
\caption{LLaMA-1-7B}
\label{subtab:llama1_7b_std}
\end{subtable}

\begin{subtable}{\textwidth}
\centering
\scalebox{0.9}{
\setlength\tabcolsep{3pt}
\begin{tabular}{p{1.5cm}|p{0.65cm}|S[table-format=2.1]*{6}{S[table-format=2.1]}}
\toprule
\textbf{Q method} & \textbf{Prec.} & \textbf{BoolQ} & \textbf{HellaSwag} & \textbf{PIQA} & \textbf{WinoGrande} & \textbf{ARC Challenge} & \textbf{ARC Easy} & \textbf{OpenBookQA}  \\
\midrule
NONE             & FP16  & 0.73 & 0.41 & 0.93 & 1.25 & 1.46 & 0.89 & 2.23 \\
\cmidrule{1-9}
NAIVE            & \centering \multirow{2}{*}{INT2} & 0.87 & 0.50 & 1.13 & 1.40 & 1.29 & 1.03 & 2.18 \\
EXAQ     &  & 0.77 & 0.43 & 0.94 & 1.27 & 1.46 & 0.90 & 2.23 \\
\cmidrule{1-9}
NAIVE            & \centering \multirow{2}{*}{INT3} & 0.77 & 0.42 & 0.95 & 1.26 & 1.46 & 0.91 & 2.23 \\
EXAQ     &  & 0.75 & 0.41 & 0.93 & 1.27 & 1.46 & 0.89 & 2.22 \\
\bottomrule
\end{tabular}
}
\caption{LLaMA-1-13B}
\label{subtab:llama1_13b_std}
\end{subtable}

\begin{subtable}{\textwidth}
\centering
\scalebox{0.9}{
\setlength\tabcolsep{3pt}
\begin{tabular}{p{1.5cm}|p{0.65cm}|S[table-format=2.1]*{6}{S[table-format=2.1]}}
\toprule
\textbf{Q method} & \textbf{Prec.} & \textbf{BoolQ} & \textbf{HellaSwag} & \textbf{PIQA} & \textbf{WinoGrande} & \textbf{ARC Challenge} & \textbf{ARC Easy} & \textbf{OpenBookQA }  \\
\midrule
NONE             & FP16  & 0.66 & 0.38 & 0.90 & 1.21 & 1.46 & 0.84 & 2.24 \\
\cmidrule{1-9}
NAIVE            & \centering \multirow{2}{*}{INT2} & 0.87 & 0.48 & 1.05 & 1.38 & 1.38 & 0.98 & 2.21 \\
EXAQ     &  & 0.69 & 0.41 & 0.91 & 1.23 & 1.46 & 0.85 & 2.24 \\
\cmidrule{1-9}
NAIVE            & \centering \multirow{2}{*}{INT3} & 0.69 & 0.39 & 0.91 & 1.21 & 1.46 & 0.84 & 2.23 \\
EXAQ     &  & 0.69 & 0.39 & 0.91 & 1.21 & 1.46 & 0.84 & 2.24 \\
\bottomrule
\end{tabular}
}
\caption{LLaMA-1-30B}
\label{subtab:llama1_30b_std}
\end{subtable}

\begin{subtable}{\textwidth}
\centering
\scalebox{0.9}{
\setlength\tabcolsep{3pt}
\begin{tabular}{p{1.5cm}|p{0.65cm}|S[table-format=2.1]*{6}{S[table-format=2.1]}}
\toprule
\textbf{Q method} & \textbf{Prec.} & \textbf{BoolQ} & \textbf{HellaSwag} & \textbf{PIQA} & \textbf{WinoGrande} & \textbf{ARC Challenge} & \textbf{ARC Easy} & \textbf{OpenBookQA }  \\
\midrule
NONE             & FP16  & 0.63 & 0.36 & 0.89 & 1.18 & 1.45 & 0.82 & 2.24 \\
\cmidrule{1-9}
NAIVE            & \centering \multirow{2}{*}{INT2} & 0.85 & 0.49 & 1.08 & 1.40 & 1.37 & 0.98 & 2.19 \\
EXAQ     &  & 0.78 & 0.40 & 0.90 & 1.19 & 1.46 & 0.84 & 2.23 \\
\cmidrule{1-9}
NAIVE            & \centering \multirow{2}{*}{INT3} & 0.66 & 0.38 & 0.91 & 1.20 & 1.46 & 0.83 & 2.23 \\
EXAQ     &  & 0.73 & 0.38 & 0.89 & 1.18 & 1.45 & 0.82 & 2.23 \\
\bottomrule
\end{tabular}
}
\caption{LLaMA-1-65B}
\label{subtab:llama1_65b_std}
\end{subtable}

\label{table:inference_accuracy_statistical_significance_llama1}
\end{table}

\subsection{Additional results}

\begin{table}[h!]
\caption{Inference accuracy evaluation for different LLaMA-2 models across various tasks}
\centering
\small

\begin{subtable}{\textwidth}
\centering
\scalebox{0.85}{
\setlength\tabcolsep{3pt}
\begin{tabular}{p{1.5cm}|p{0.65cm}|S[table-format=2.1]*{6}{S[table-format=2.1]}|>{\columncolor{lightgray1}}S[table-format=2.1]}
\toprule
\textbf{Q method} & \textbf{Prec.} & \textbf{BoolQ} & \textbf{HellaSwag} & \textbf{PIQA} & \textbf{WinoGrande} & \textbf{ARC Challenge} & \textbf{ARC Easy} & \textbf{OpenBookQA} & \textbf{avg score} \\
\midrule
NONE               & \centering BF16 & 77.9 & 76.0 & 78.9 & 69.1 & 46.2 & 74.8 & 44.2 & 66.7 \\
\cmidrule{1-10}
NAIVE            & \centering \multirow{2}{*}{INT2} & 58.6 & 33.5 & 61.4 & 51.3 & 25.2 & 40.0 & 29.6 & 42.8 \\
\textbf{EXAQ}     &  & \textbf{73.7} & \textbf{74.4} & \textbf{78.0} & \textbf{68.4} & \textbf{44.5} & \textbf{72.3} & \textbf{42.2} & \textbf{64.8} \\
\cmidrule{1-10}
NAIVE            & \centering \multirow{2}{*}{INT3} & 69.9 & 72.5 & 77.6 & 66.9 & 43.4 & 70.2 & 42.8 & 63.3 \\
\textbf{EXAQ}     & & \textbf{75.9} & \textbf{75.5} & \textbf{78.9} & \textbf{68.8} & \textbf{46.4} & \textbf{74.8} & \textbf{44.0} & \textbf{66.3}  \\
\bottomrule
\end{tabular}
}
\caption{LLaMA-2-7B}
\label{subtab:llama2_7b}
\end{subtable}


\begin{subtable}{\textwidth}
\centering
\scalebox{0.85}{
\setlength\tabcolsep{3pt}
\begin{tabular}{p{1.5cm}|p{0.65cm}|S[table-format=2.1]*{6}{S[table-format=2.1]}|>{\columncolor{lightgray1}}S[table-format=2.1]}
\toprule
\textbf{Q method} & \textbf{Prec.} & \textbf{BoolQ} & \textbf{HellaSwag} & \textbf{PIQA} & \textbf{WinoGrande} & \textbf{ARC Challenge} & \textbf{ARC Easy} & \textbf{OpenBookQA} & \textbf{avg score} \\
\midrule
NONE               & \centering BF16 & 80.7 & 79.3 & 80.6 & 72.5 & 49.4 & 77.3 & 45.6 & 69.3 \\
\cmidrule{1-10}
NAIVE            & \centering \multirow{2}{*}{INT2} & 54.9 & 35.6 & 58.4 & 51.2 & 24.9 & 41.4 & 33.4 & 42.8 \\
\textbf{EXAQ}     &  & \textbf{77.5} & \textbf{77.7} & \textbf{79.4} & \textbf{70.0} & \textbf{48.5} & \textbf{76.9} & \textbf{46.6} & \textbf{68.1} \\
\cmidrule{1-10}
NAIVE            & \centering \multirow{2}{*}{INT3} & 72.1 & 77.0 & 79.1 & 70.2 & 48.2 & 74.8 & \textbf{44.8} & 66.6  \\
\textbf{EXAQ}   &  & \textbf{79.7} & \textbf{78.9} & \textbf{80.0} & \textbf{71.7} & \textbf{48.5} & \textbf{77.5} & 44.4 & \textbf{68.7} \\
\bottomrule
\end{tabular}
}
\caption{LLaMA-2-13B}
\label{subtab:llama2_13b}
\end{subtable}


\begin{subtable}{\textwidth}
\centering
\scalebox{0.85}{
\setlength\tabcolsep{3pt}
\begin{tabular}{p{1.5cm}|p{0.65cm}|S[table-format=2.1]*{6}{S[table-format=2.1]}|>{\columncolor{lightgray1}}S[table-format=2.1]}
\toprule
\textbf{Q method} & \textbf{Prec.} & \textbf{BoolQ} & \textbf{HellaSwag} & \textbf{PIQA} & \textbf{WinoGrande} & \textbf{ARC Challenge} & \textbf{ARC Easy} & \textbf{OpenBookQA } & \textbf{avg score} \\
\midrule
NONE               & \centering BF16 & 83.6 & 83.8 & 82.8 & 77.9 & 57.5 & 80.9 & 48.4 & 73.6 \\
\cmidrule{1-10}
NAIVE            & \centering \multirow{2}{*}{INT2} & 51.1 & 46.5 & 69.3 & 55.5 & 30.3 & 64.9 & 45.4 & 51.9 \\
\textbf{EXAQ}     &  & \textbf{74.8} & \textbf{73.3} & \textbf{82.4} & \textbf{76.2} & \textbf{54.8} & \textbf{79.4} & \textbf{48.2} & \textbf{69.9} \\
\cmidrule{1-10}
NAIVE            & \centering \multirow{2}{*}{INT3} &  \textbf{79.0} & \textbf{83.2} & 82.6 & 75.7 & 57.0 & \textbf{80.7} & 48.6 & \textbf{72.4}  \\
\textbf{EXAQ}     &  & 77.9 & 78.9 & \textbf{82.9} & \textbf{77.0} & 56.9 & 80.0 & \textbf{49.0} & 71.8  \\
\bottomrule
\end{tabular}
}
\caption{LLaMA-2-70B}
\label{subtab:llama2_70b}
\end{subtable}

\label{table:inference_accuracy_llama2}
\end{table}

\begin{table}[h!]
\caption{The standard deviation $(\sigma)$ over multiple runs of LLaMA-2 models. The mean values of these runs are presented above in Table.\ref{table:inference_accuracy_llama2} in the appendix.}
\centering
\small

\begin{subtable}{\textwidth}
\centering
\scalebox{0.85}{
\setlength\tabcolsep{3pt}
\begin{tabular}{p{1.5cm}|p{0.65cm}|S[table-format=2.1]*{6}{S[table-format=2.1]}}
\toprule
\textbf{Q method} & \textbf{Prec.} & \textbf{BoolQ} & \textbf{HellaSwag} & \textbf{PIQA} & \textbf{WinoGrande} & \textbf{ARC Challenge} & \textbf{ARC Easy} & \textbf{OpenBookQA} \\
\midrule
NONE             & FP16 & 0.73 &  0.43 & 0.95 & 1.30 & 1.46 & 0.89 & 2.22 \\
\cmidrule{1-9}
NAIVE            & \centering \multirow{2}{*}{INT2} & 0.86   & 0.47     & 1.14 & 1.40      & 1.27        & 1.01    & 2.04 \\
\textbf{EXAQ}     &  & 0.77   & 0.44     & 0.97 & 1.31      & 1.45        & 0.92    & 2.21  \\
\cmidrule{1-9}
NAIVE            & \centering \multirow{2}{*}{INT3} & 0.80   & 0.45     & 0.97 & 1.32      & 1.45        & 0.94    & 2.21  \\
\textbf{EXAQ}     & & 0.75   & 0.43     & 0.95 & 1.30      & 1.46        & 0.89    & 2.22   \\
\bottomrule
\end{tabular}
}
\caption{LLaMA-2-7B}
\label{subtab:llama2_7b_std}
\end{subtable}

\begin{subtable}{\textwidth}
\centering
\scalebox{0.85}{
\setlength\tabcolsep{3pt}
\begin{tabular}{p{1.5cm}|p{0.65cm}|S[table-format=2.1]*{6}{S[table-format=2.1]}}
\toprule
\textbf{Q method} & \textbf{Prec.} & \textbf{BoolQ} & \textbf{HellaSwag} & \textbf{PIQA} & \textbf{WinoGrande} & \textbf{ARC Challenge} & \textbf{ARC Easy} & \textbf{OpenBookQA}  \\
\midrule
NONE             & FP16  & 0.69 & 0.40 & 0.92 & 1.26 &  1.46 & 0.86 & 2.23 \\
\cmidrule{1-9}
NAIVE            & \centering \multirow{2}{*}{INT2} & 0.87   & 0.48     & 1.15 & 1.40      & 1.26        & 1.01    & 2.11  \\
\textbf{EXAQ}     &  & 0.73   & 0.42     & 0.94 & 1.29      & 1.46        & 0.87    & 2.23   \\
\cmidrule{1-9}
NAIVE            & \centering \multirow{2}{*}{INT3} & 0.78   & 0.42     & 0.95 & 1.29      & 1.46        & 0.89    & 2.23   \\
\textbf{EXAQ}   &  & 0.70   & 0.41     & 0.93 & 1.27      & 1.46        & 0.86    & 2.22   \\
\bottomrule
\end{tabular}
}
\caption{LLaMA-2-13B}
\label{subtab:llama2_13b_std}
\end{subtable}

\begin{subtable}{\textwidth}
\centering
\scalebox{0.85}{
\setlength\tabcolsep{3pt}
\begin{tabular}{p{1.5cm}|p{0.65cm}|S[table-format=2.1]*{6}{S[table-format=2.1]}}
\toprule
\textbf{Q method} & \textbf{Prec.} & \textbf{BoolQ} & \textbf{HellaSwag} & \textbf{PIQA} & \textbf{WinoGrande} & \textbf{ARC Challenge} & \textbf{ARC Easy} & \textbf{OpenBookQA }  \\
\midrule
NONE             & FP16  & 0.65 & 0.37 & 0.88 & 1.17 & 1.44 & 0.81 & 2.24      \\
\cmidrule{1-9}
NAIVE            & \centering \multirow{2}{*}{INT2} & 0.87   & 0.50     & 1.08 & 1.40      & 1.34        & 0.98    & 2.23   \\
\textbf{EXAQ}     &  & 0.76   & 0.44     & 0.89 & 1.20      & 1.45        & 0.83    & 2.24   \\
\cmidrule{1-9}
NAIVE            & \centering \multirow{2}{*}{INT3} &  0.71   & 0.37     & 0.88 & 1.21      & 1.45        & 0.81    & 2.24   \\
\textbf{EXAQ}     &  & 0.73   & 0.41     & 0.88 & 1.18      & 1.45        & 0.82    & 2.24    \\
\bottomrule
\end{tabular}
}
\caption{LLaMA-2-70B}
\label{subtab:llama2_70b_std}
\end{subtable}

\label{table:inference_accuracy_statistical_significance_llama2}
\end{table}

\end{document}